\documentclass[10pt,twocolumn,letterpaper]{article}

\usepackage[pagenumbers,preprint]{iccv} 

\definecolor{iccvblue}{rgb}{0.21,0.49,0.74}
\usepackage[pagebackref,breaklinks,colorlinks,allcolors=iccvblue]{hyperref}
\usepackage{pgf-pie} 
\usepackage{pgfplots}

\usepackage{graphicx}
\DeclareGraphicsExtensions{.pdf,.png,.jpg}

\def\paperID{8414} 
\def\confName{ICCV}
\def\confYear{2025}

\title{DiffAD: A Unified Diffusion Modeling Approach for Autonomous Driving}

\author{
Tao Wang$^{1}$, Cong Zhang$^{1}$, Xingguang Qu$^{2}$, Kun Li$^{1}$, Weiwei Liu$^{1}$, Chang Huang$^{1}$\\
$^{1}$Carizon $^{2}$Beihang University\\
}

\begin{document}
\maketitle
\begin{abstract}
End-to-end autonomous driving (E2E-AD) has rapidly emerged as a promising approach toward achieving full autonomy. However, existing E2E-AD systems typically adopt a traditional multi-task framework, addressing perception, prediction, and planning tasks through separate task-specific heads. Despite being trained in a fully differentiable manner, they still encounter issues with task coordination and the system complexity remains high. In this work, we introduce DiffAD—a novel diffusion probabilistic model that redefines autonomous driving as a conditional image generation task. By rasterizing heterogeneous targets onto a unified bird’s-eye view (BEV) and modeling their latent distribution, DiffAD unifies various driving objectives and jointly optimizes all driving tasks in a single framework, significantly reduces system complexity and harmonizes task coordination. The reverse process iteratively refine the generated BEV image, resulting in more robust and realistic driving behaviors. Closed-loop evaluations in Carla demonstrate the superiority of the proposed method, achieving a new state-of-the-art Success Rate and Driving Score. Code and models are available at \href{https://github.com/wantsu/DiffAD#}{\texttt{https://github.com/wantsu/DiffAD}}.

\end{abstract}    
\section{Introduction}
\label{sec:intro}
Achieving full autonomy in driving requires not only a deep understanding of complex scenes but also effective interaction with dynamic environments and comprehensive learning of driving behaviors. Traditional autonomous driving systems are built upon a modular architecture, where perception, prediction, and planning are developed independently and then integrated into the onboard system. While this design offers interpretability and facilitates debugging, the separate optimization objectives across modules often lead to information loss and error accumulation.

\begin{figure}[t]
  \centering
   \includegraphics[width=1.\linewidth]{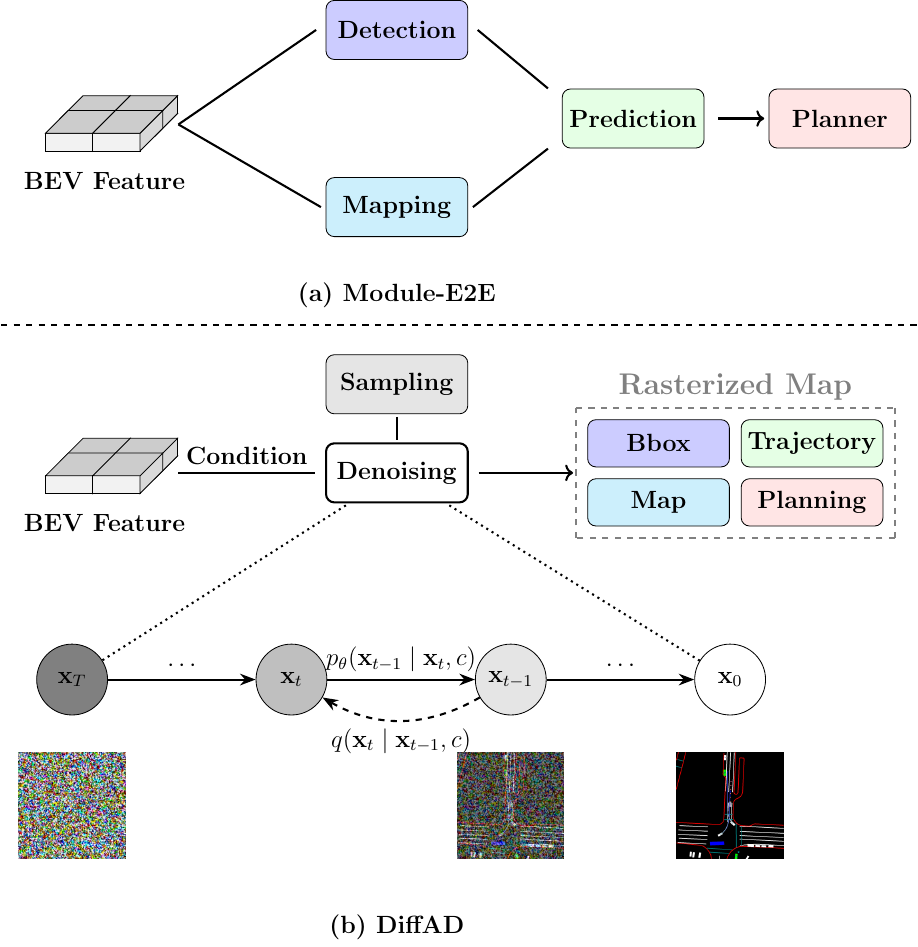}

   \caption{Paradigm overview. (a) Module-E2E adopts sequential pipelines where multi-task heads is optimized in a differentiable manner. (b) DiffAD (ours) integrates all components into a single denoising head and treat E2E-AD as a conditional image generation task, resulting a fully end-to-end joint optimization of all driving tasks.}
   \label{fig:module-e2e}
\end{figure}

Recent end-to-end autonomous driving (E2E-AD) approaches (e.g., \cite{mp3,uniad,vad}) have attempted to overcome these limitations by enabling joint, fully differentiable training of all components, as illustrated in \cref{fig:module-e2e}(a). However, several critical issues remain:

\begin{enumerate}
\item \textbf{Sub-optimal Optimization}: Methods like UniAD \cite{uniad} and VAD \cite{vad} still rely on sequential pipelines, where the planning stage depends on the outputs of preceding modules. This dependency can amplify errors throughout the system.
\item \textbf{Inefficient Query Modeling}: Current query-based methods (e.g., \cite{vad, uniad}) deploy thousands of learnable queries to capture potential traffic elements. This approach leads to an inefficient allocation of computational resources, with a disproportionate focus on upstream auxiliary tasks rather than the core planning module. For instance, in VAD, the perception task consumes 34.6\% of the total runtime, while the planning module accounts for only 5.7\%.
\item \textbf{Complexity in Coordination}: With each task head optimized independently using distinct objective functions—and with targets varying in shape and semantic meaning—the overall system becomes fragmented and difficult to train cohesively \cite{e2esurvey}.
\end{enumerate}

To address these limitations, we propose a novel paradigm, \textbf{DiffAD}, which unifies the optimization of all driving tasks within a single model, as depicted in \cref{fig:module-e2e}(b). Specifically, we rasterize heterogeneous targets from perception, prediction, and planning onto a unified bird's-eye view (BEV) space, thereby recasting the autonomous driving problem as one of conditional image generation. A denoising diffusion probabilistic model is employed to learn the distribution of the BEV image conditioned on surrounding views. This approach not only enables the simultaneous optimization of all tasks—thereby mitigating error propagation—but also replaces inefficient vector-based query methods with a computationally efficient generative modeling strategy in latent space using a shared decoding head. Moreover, by focusing solely on noise prediction without the need for multiple loss functions or complex bipartite matching, our method significantly simplifies the overall training procedure.

In summary, DiffAD overcomes the limitations of existing query-based, sequential methods by unifying tasks into a single, end-to-end framework that enhances coordination, reduces error propagation, and allocates computational resources more efficiently towards safe and effective planning. The main contributions of this paper are summarized as follows:
\begin{itemize}
    \item We introduce an end-to-end paradigm for autonomous driving that leverages a unified, fully rasterized BEV representation to integrate diverse driving tasks into a single model.

    \item We reformulate driving tasks as a conditional image generation problem and present \textbf{DiffAD}, a diffusion model that learns the latent distribution of BEV images conditioned on surrounding views. Additionally, we propose a data-driven method to extract vectorized planning trajectories from the generated BEV images.

    \item We demonstrate that DiffAD achieves state-of-the-art performance in end-to-end planning, significantly outperforming previous methods in closed-loop evaluations.
\end{itemize}


\section{Related Work}
\label{sec:related}
We cover previous works on End-to-End autonomous driving, Driving VLM and Diffusion Model for the downstream tasks of end-to-end driving.

\paragraph{End-to-End Autonomous Driving.} 
Traditionally, autonomous driving systems are composed of separate modules for detection \cite{lss, bevformer, detr3d, petr}, mapping \cite{hdmapnet, vectormapnet, maptr}, prediction \cite{vectornet, densetnt, scenetransformer}, and planning. While this modular design facilitates task-specific optimizations, it often leads to information loss and error accumulation when integrating these components, resulting in suboptimal planning decisions. End-to-end (E2E) approaches seek to address these limitations by unifying all tasks within a fully differentiable framework, thereby enabling planning-oriented optimization. For instance, methods such as UniAD \cite{uniad} and VAD \cite{vad} utilize query-based architectures to transfer information from perception to planning. Paradrive \cite{paradrive} employs a parallel learning pipeline, directly optimizing all tasks from dense BEV features, while SparseAD \cite{sparsead} adopts a sparse query-based strategy to bypass the inefficiencies of dense feature construction. However, these approaches rely on multi-head instance query modeling, which can introduce coordination challenges across tasks and lead to an inefficient allocation of computational resources toward auxiliary tasks rather than core planning. In contrast, our method, DiffAD, redefines autonomous driving as a conditional image generation task in a unified bird’s-eye view space. By jointly optimizing perception, prediction, and planning within a single diffusion framework, DiffAD streamlines the overall optimization process, mitigates error propagation, and prioritizes safe and coherent planning outcomes.

\paragraph{Vision-Language Models (VLMs) for Driving.}

Recent works \cite{drivegpt4, talk2bev, drivelm, drivevlm} have explored applying large Vision-Language Models (VLMs) to autonomous driving. These models leverage the reasoning capabilities of large language models (LLMs) to provide natural language explanations for driving decisions, enhancing interpretability and generalization. However, deploying such models on edge devices for real-time inference remains challenging, and LLMs are prone to generating inaccurate or misleading outputs (hallucinations), which could compromise safety in autonomous driving.

\paragraph{Diffusion Model for Autonomous Driving.}
Denoising diffusion models \cite{ddpm, score-based} have recently emerged as a powerful class of generative models, achieving state-of-the-art results in diverse applications such as image generation \cite{dmppp, diffusionbeatsgans, zhang2023adding, Saharia2024text2image}, video generation \cite{stablevideodiffusion, videodm}, and image editing \cite{imagic}. In autonomous driving, several works have begun exploring their potential. For example, DiffBEV \cite{diffbev} leverages conditional diffusion models to generate a refined BEV representation with reduced noise, while PolyDiffuse \cite{polydiffuse} employs guided diffusion to reconstruct polygonal shapes for mapping, and MotionDiffuser \cite{motiondiffuser} utilizes diffusion-based representations to predict multi-agent trajectories. Similarly, DiffusionDrive \cite{diffusiondrive} adopts a truncated diffusion process to capture multi-modal trajectory distributions. However, these methods primarily focus on refining specific components—improving BEV perception or trajectory prediction—and are built upon existing sequential multi-task frameworks. In contrast, our work is the first to formulate holistic end-to-end autonomous driving as a conditional image generation problem. By jointly learning perception, prediction, and planning within a unified diffusion framework, our approach significantly simplifies system architecture and enhances task coordination. This fundamental shift in problem formulation not only streamlines the overall optimization process but also leverages the inherent denoising capabilities of diffusion models to produce more robust and realistic driving decisions.
\section{Preliminary: Diffusion Models}
\label{sec:preliminary}

Diffusion models, also known as score-based generative models \cite{ddpm, dulnt, score-based}, progressively inject noise into the data during the forward (diffusion) process and generate data from noise through the reverse (denoising) process. This section provides key preliminary knowledge about denoising diffusion probabilistic models (DDPM), laying the groundwork for DiffAD.

\paragraph{Forward Process.} 
DDPM considers a diffusion process that transforms data $\mathbf{x}_0 \sim q(\mathbf{x}_0)$ into Gaussian noise: 
\begin{equation} q(\mathbf{x}_t|\mathbf{x}_{t-1}) := \mathcal{N}(\mathbf{x}_t; \sqrt{1 - \beta_t} \mathbf{x}_{t-1}, \beta_t \mathbf{I}), \end{equation} 
for $t = 1, \dots, T$, where $\mathbf{x}_t$ represents the latent variable at time $t$. The noise schedule ${\beta_t}$ can either be constant \cite{ddpm, dulnt} or learned via reparameterization \cite{improveddm}. This forward process can be expressed in closed form for any $t$: \begin{equation} q(\mathbf{x}_t|\mathbf{x}_0) = \mathcal{N}(\mathbf{x}_t; \sqrt{\bar{\alpha}_t} \mathbf{x}_0, (1 - \bar{\alpha}_t)\mathbf{I}), \end{equation}
where $\bar{\alpha}_0 = 1$, $\bar{\alpha}_t := \prod_{s=1}^t \alpha_s$, and $\alpha_t := 1 - \beta_t$. The latent variable $\mathbf{x}_t$ is then a linear combination of $\mathbf{x}_0$ and noise: \begin{equation} \mathbf{x}_t = \sqrt{\bar{\alpha}_t}\mathbf{x}_0 + \sqrt{1 - \bar{\alpha}_t}\mathbf{\epsilon}, \quad \mathbf{\epsilon} \sim \mathcal{N}(\mathbf{0}, \mathbf{I}). \end{equation}

\paragraph{Reverse Process.}
Diffusion models are trained to learn the reverse process $p_\theta(x_{t-1}|x_t) = \mathcal{N}(\mu_\theta(x_t), \Sigma_\theta(x_t))$, where neural networks predict the parameters of $p_\theta$. The DDPM uses a noise prediction (denoising) network $\mathbf{\epsilon}_\theta(\mathbf{x}_t, t)$ to connect the process with denoising score matching and Langevin dynamics \cite{song2020gmeg, vincent2011}. The sampling step of the reverse process is derived as:
\begin{equation}
    \mathbf{x}_{t-1} = \frac{1}{\sqrt{\alpha_t}} \left[ \mathbf{x}_t - \frac{1 - \alpha_t}{\sqrt{1 - \bar{\alpha}_t}} \mathbf{\epsilon}_\theta(\mathbf{x}_t, t) \right] + \sigma_t \mathbf{z}, \quad \mathbf{z} \sim \mathcal{N}(\mathbf{0}, \mathbf{I}).
\end{equation}
where $\sigma_t^2$ is set to either $\beta_t$ or $\frac{1-\bar{\alpha}_{t-1}}{1-\bar{\alpha}_t}\beta_t$. The final training objective is a reweighted variational lower bound:
\begin{equation}
    L_\text{simple}(\theta) := \mathbb{E}_{\mathbf{x}_0, t, \mathbf{\epsilon}} \left[\|\mathbf{\epsilon} - \mathbf{\epsilon}_\theta(\sqrt{\bar{\alpha}_t}\mathbf{x}_0 + \sqrt{1 - \bar{\alpha}_t} \mathbf{\epsilon}, t)\|^2 \right].
\end{equation}


\section{DiffAD}
\label{sec:method}

\begin{figure*}[t] 
  \centering
  \includegraphics[height=0.32\textheight]{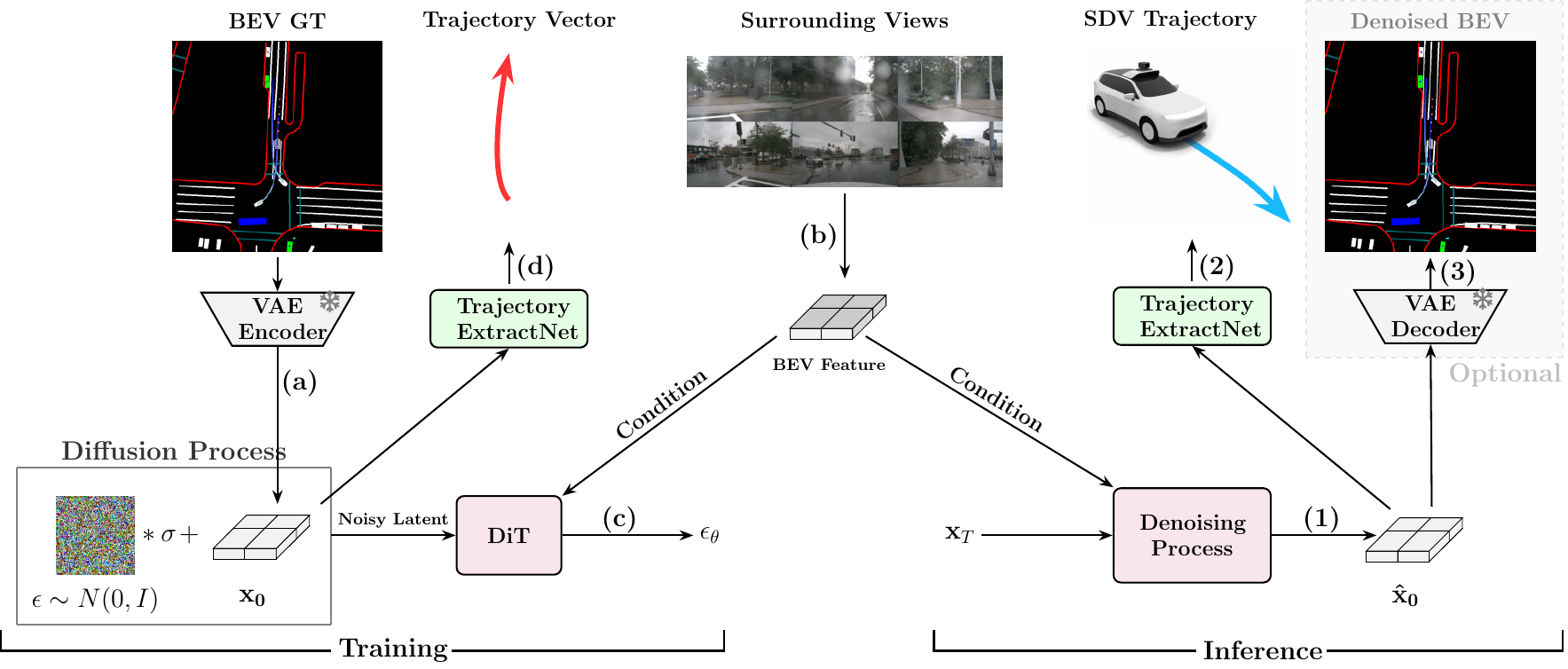} 
  \caption{Pipeline of DiffAD. Training Process: (a) DiffAD rasterizes perception, prediction, and planning targets onto a BEV image, which is encoded into a latent space $\mathbf{x_0}$ using a VAE. (b) Surrounding images are transformed into BEV feature. (c) A diffusion model predicts noise $\mathbf{\epsilon_\theta}$ from the noisy latent BEV image, and (d) a trajectory extraction network (TEN) learns to recover the ego trajectory from the latent BEV image.
   Inference Process: (1) DiffAD generates a denoised latent BEV image $\mathbf{\hat{x}_0}$ from pure Gaussian noise $\mathbf{x}_T$, conditioned on BEV feature, (2) extracts the ego trajectory via TEN, and (3) decodes the latent BEV image for interpretation.}
  \label{fig:architecture}
\end{figure*}

\paragraph{Overview.}
As illustrated in \cref{fig:architecture}, DiffAD consists of three main components: a Latent Diffusion Model, a BEV Feature Generator, and a Trajectory Extraction Network (TEN).
\\\textbf{Training Process:}
\begin{enumerate}
    \item \textbf{Rasterization and Latent Space Encoding:} DiffAD begins by rasterizing the perception, prediction, and planning targets into a BEV image. An off-the-shelf VAE Encoder is then used to compress the BEV image into a latent space for dimensionality reduction.
    
    \item \textbf{Feature Extraction and Transformation:} Surrounding view images are fed into a feature extractor, which transforms the resulting perspective-view features into unified BEV features.

    \item \textbf{Diffusion Model for Noise Prediction:} 
    Gaussian noise is added to the latent BEV image to obtain noisy latent. A diffusion model is trained to predict the noise from the noisy latent representation, conditioned on the BEV features.

    \item \textbf{Trajectory Extraction:} A query-based TEN is trained to recover the vectorized trajectory of the ego agent from the latent BEV image.

\end{enumerate}
\textbf{Inference Process:}

\begin{enumerate}
    \item \textbf{Conditional Denoising:} DiffAD first generates a denoised latent BEV image from pure Gaussian noise, conditioned on the BEV features.
    
    \item \textbf{Planning Extraction:} The TEN then extracts the planned trajectory of the ego agent from the latent BEV image.
    
    \item \textbf{Decoding BEV:} By decoding the latent BEV image back into pixel space, we can obtain the predicted BEV image for interpretation and debugging.
\end{enumerate}

\subsection{Rasterized BEV Representation}
Perceiving surrounding traffic agents and map elements is essential for understanding the driving scene, while predicting agents' trajectories is crucial for making safe driving decisions. DiffAD utilizes a rasterized BEV representation to unify the heterogeneous targets of driving tasks—such as bounding boxes, lane elements, agent trajectories, and ego vehicle planning.

Specifically, we rasterize the bounding boxes and map elements onto an RGB canvas for the perception task, denoted as $m_{perc} \in \mathbb{R}^{3 \times H \times W}$, where different semantic elements are represented using distinct colors. For the trajectory prediction task, agents' trajectories are drawn onto second RGB canvas, $m_{pre} \in \mathbb{R}^{3 \times H \times W}$. Finally, the ego vehicle's future trajectory is rasterized on third RGB canvas for the planning task, $m_{plan} \in \mathbb{R}^{3 \times H \times W}$. The color of the trajectories is interpolated over time to represent the temporal relationship between points. 

This unified BEV representation allows the diffusion model to simultaneously learn the tasks of perception, prediction, and planning. Moreover, it enables reasoning about the physical relationships and social interactions between the ego vehicle and its surroundings, leading to scene-level consistency results across all tasks.

\subsection{Denoising Diffusion Learning}
Following the Latent Diffusion Model (LDM) framework \cite{ldm}, we utilize a pretrained VAE to compress the rasterized BEV images into a low-dimensional latent space. The latent representations of perception, prediction, and planning are then concatenated along the channel dimension to construct the latent BEV image $z_{bev} \in \mathbb{R}^{c' \times h' \times w'}$.

\begin{equation}
\begin{aligned}
[z_{perc}, z_{pre}, z_{plan}] &= \text{encoder}([m_{perc}, m_{pre}, m_{plan}]), \\
z_{bev} &= \text{concat}([z_{perc}, z_{pre}, z_{plan}])
\end{aligned}
\end{equation}
Next, noise $\epsilon$ is added via the diffusion process to produce a noisy latent image $\{z_t\}_{t=0}^T$ at each timestep $t$, where $z_0 = z$. The noisy latent image is divided into tokens and passed through multiple layers of DiT \cite{dit}, with an MLP layer used at the end to predict the noise $\epsilon_\theta$.

\paragraph{Conditional Denoising.} 
DiffAD utilizes multi-view images and driving commands \cite{uniad, vad} as conditions to guide the denoising process. For the conditional guidance mechanism, we adopt Adaptive Layer Normalization (AdaLN) with zero-initialization \cite{dit} due to its effectiveness and efficiency. Specifically, we employ BEVFormer \cite{bevformer} to convert multi-view images into a BEV feature map $x_{bev} \in \mathbb{R}^{c_{bev} \times h' \times w'}$, then the BEV feature $x_{bev}$ is tokenized and combined with the timestep embedding and driving command embedding $x_{cmd}$ as input to AdaLN.

\begin{equation}
\begin{aligned}
\text{cond} &= t_{emb} + x_{bev} + x_{cmd}, \\
z_t &= \text{AdaLN}(z_t, \text{cond})
\end{aligned}
\end{equation}\\
\textbf{Temporal-Consistency.} Planning is fundamentally a sequential decision-making task, where the agent must make decisions based on its current status and the dynamics of the environment. To capture temporal information, we adopt ConvLSTM to fuse historical BEV features. However, fusing BEV features alone is insufficient to ensure consistent planning over time. To address this problem, We introduce a \textbf{Action-Guidance} mechanisms, where we take an assumption that the current decision depends not only on the current observation but also on the last action. Thus, the joint distribution can be modeled as $\prod_{t=1}^{T} q(a_t \mid s_t, a_{t-1})$, where \( s_t \) represents the agent's state at time \( t \), \( a_t \) represents the action taken at time \( t \). For the implementation, we condition the current output on the previous latent BEV image \( z_{bev}^{t-1} \) \cite{teacherforcing} to regularize the current decision. This approach, however, could lead to a shortcut, where the network over-relies on the previous latent BEV image and neglects the current observation. To mitigate this issue, we introduce a dropout regularization on the previous latent BEV image tokens with a probability of 0.5. The final conditional guidance is formulated as follows:

\begin{equation}
\begin{aligned}
\text{cond} &= t_{emb} + x_{bev} + x_{cmd} + \sigma_{p=0.5}(z_{bev}^{t-1}),
\end{aligned}
\end{equation}
where \( \sigma_{p=0.5} \) represents the random dropout operation.

\subsection{Trajectory Extraction Network}
To obtain a vectorized trajectory for ego vehicle control, we need to recover the trajectory from the latent space. A straightforward approach would be to decode the latent BEV image back into pixel space and apply a rule-based post-processing method. However, to improve generalization and robustness, we opt for a data-driven approach. 

Specifically, we design a query-based transformer network\cite{detr} to extract the trajectory from the latent BEV image.
First, the latent BEV image $z_{bev} \in \mathbb{R}^{12\times h'\times w'}$ is split into a sequence of tokens $X \in \mathbb{R}^{L\times D}$ through an embedding layer $f_{emb}$. A learnable query $Q_{\text{ego}} \in \mathbb{R}^{T\times D}$ interacts with the tokenized sequence via a series of transformer layers. Finally, a single MLP decodes the learned query into the predicted trajectory $\hat{\tau} \in \mathbb{R}^{T\times 2}$. The process is summarized as follows:

\begin{equation}
\begin{aligned}
X &= f_{emb}(z_{bev}), \\
Q'_{\text{ego}} &= \text{Transformer}(Q=Q_{\text{ego}}, K=X, V=X), \\
\hat{\tau} &= \text{MLP}(Q'_{\text{ego}}).
\end{aligned}
\end{equation}

\subsection{End-to-End Learning}
DiffAD is fully end-to-end trainable, based on the rasterized BEV representation and the diffusion model. Unlike traditional Module-E2E approaches, which involve multiple loss functions for different driving tasks, our system simplifies the optimization by using a unified loss function: the noise regression loss for denoising, and a trajectory extraction loss for the vectorized trajectory.

\begin{equation}
  \mathcal{L}  = \mathcal{L}_{\text{denoising}} + \mathcal{L}_{\text{extraction}}
  \label{eq:important}
\end{equation}

\paragraph{Denoising Loss.}
We use the standard mean squared error (MSE) loss to optimize the diffusion model, ensuring it can accurately recover the noise from the noisy latent BEV image.
\begin{equation}
  \mathcal{L}_{\text{denoising}}  = \frac{1}{N} \sum \left \| \epsilon_\theta - \epsilon  \right \|^2
  \label{eq:important}
\end{equation}

\paragraph{Trajectory Extraction Loss.}
The trajectory extraction loss is also based on MSE, applied between the predicted trajectory $\hat{\tau}$ and the ground truth ego trajectory $\tau$. This loss ensures that the network can accurately recover the vectorized trajectory from the latent BEV image.
\begin{equation}
  \mathcal{L}_{\text{extraction}}  = \frac{1}{N} \sum \left \| \hat{\tau} - \tau  \right \|^2
  \label{eq:important}
\end{equation}

\section{Experiments}
\label{sec:experiments}

\subsection{Datasets}
Open-loop evaluations have been reported as insufficient for E2E models \cite{bench2drive, is_ego_status}. To address this, we use the Bench2Drive dataset for training and closed-loop evaluation in the CARLA simulator\cite{carla}. Bench2Drive offers three data subsets: mini (10 clips for debugging), base (1,000 clips), and full (10,000 clips for large-scale studies). Following the methodology of \cite{bench2drive}, we use the base subset for training.

\begin{table*}[h]
    \centering
    \setlength{\tabcolsep}{2pt}
    \caption{Multi-Ability and Overall Results of E2E-AD Methods in Bench2Drive. * denotes expert feature distillation.}
    \begin{tabular}{l c c c c c c c c c}
        \toprule
        \textbf{Method} & \multicolumn{6}{c}{\textbf{Multi-Ability} (\%) $\uparrow$} & \multicolumn{2}{c}{\textbf{Overall}} \\
        \cmidrule(lr){2-7} \cmidrule(lr){8-9}
        & Merging & Overtaking & Emergency Brake & Give Way & Traffic Sign & Mean & Driving Score $\uparrow$ & Success Rate $\uparrow$ \\
        \midrule
        AD-MLP        & 0.00   & 0.00   & 0.00   & 0.00   & 4.35   & 0.87  & 18.05  & 0.00   \\
        UniAD-Tiny    & 8.89   & 9.33   & 20.00  & 20.00  & 15.43  & 14.73  & 40.73 & 13.18   \\
        UniAD-Base    & 14.10  & 17.78  & 21.67  & 10.00  & 14.21  & 15.55  & 45.81 & 16.36   \\
        VAD           & 8.11  & 24.44   & 18.64  & 20.00  & 19.15  & 18.07  & 42.35 & 15.00  \\
        \midrule
        TCP*          & 16.18   &20.00  & 20.00   & 10.00  & 6.99   & 14.63  & 40.70 & 15.00   \\
        TCP-ctrl*     & 10.29   &4.44   & 10.00   & 10.00   & 6.45  & 8.23   & 30.47 & 7.27   \\
        TCP-traj*     & 8.89    &24.29  & \textbf{51.67}   & 40.00  & 46.28  & 34.22  & 59.90 & 30.00  \\
        ThinkTwice*   & 27.38   &18.42  & 35.82   & \textbf{50.00}  & 54.23   & 37.17 & 62.44 & 31.23  \\
        DriveAdapter* & 28.82   &26.38  & 48.76   & \textbf{50.00}  &\textbf{56.43}   & \textbf{42.08} & 64.22 & 33.08 \\
        \midrule
        DiffAD (ours)     & \textbf{30.00}  & \textbf{35.55}  & 46.66  & 40.00  & 46.32  & 38.79  & \textbf{67.92} & \textbf{38.64}  \\
        \bottomrule
    \end{tabular}
    \label{tab:combined_results}
\end{table*}

\subsection{Metrics}

\begin{itemize}
    \item \textbf{Success Rate (SR)\cite{bench2drive}:}
    This metric calculates the proportion of routes successfully completed without traffic violations within the allotted time.
    \item \textbf{Driving Score (DS)\cite{bench2drive}:}
    This metric considers both route completion and penalties for infractions.
    \item \textbf{FID:}
    We use the Frechet Inception Distance (FID) \cite{fid} to assess scaling performance, which is a standard metric for evaluating generative models of images.
    \end{itemize}

\subsection{Baselines}

\begin{itemize}
    \item \textbf{UniAD \cite{uniad}:} A classic module-based E2E approach that employs a query-based architecture to explicitly link perception, prediction, and planning tasks.

    \item \textbf{VAD \cite{vad}:} Another module-based E2E method, which enhances computational efficiency by utilizing Transformer Queries with a vectorized scene representation.

    \item \textbf{AD-MLP \cite{admlp}:} A baseline model that predicts future trajectories by simply feeding the ego vehicle's historical states into an MLP.

    \item \textbf{TCP \cite{tcp}:} A simple yet effective baseline, using only the front cameras and ego state to predict both trajectories and control commands.

    \item \textbf{ThinkTwice \cite{thinktwice}:} A method that promotes a coarse-to-fine framework, refining planning routes iteratively and leveraging expert feature distillation.

    \item \textbf{DriveAdapter \cite{driveadapter}:} The top-performing method on the Bench2Drive leaderboard, which fully utilizes expert feature distillation to enhance performance by decoupling perception and planning.
\end{itemize}

\subsection{Implementation Details}

\paragraph{Training.}
We use an off-the-shelf pre-trained variational autoencoder (VAE) model \cite{vae} from Stable Diffusion\cite{ldm}. The VAE encoder has a downsample factor of 8. Across all experiments in this section, our diffusion models operate in latent space. We retain diffusion hyperparameters from DiT \cite{dit}. 
To facilitate the learning process, we start with single image learning in the first stage for perception parts, i.e., detection and mapping, while prediction and planning BEV images are padding with zero. Then train the model jointly with all perception, prediction and planning parts in temporal setting.

\paragraph{Inference.}
We utilize the DDIM-10 sampler \cite{ddim} for inference and employ official evaluation tools \cite{bench2drive} to compute closed-loop metrics. For vehicle control, we adopt the officially provided PID controller.

\subsection{Main Results.}
The overall results in \cref{tab:combined_results} shows that DiffAD significantly outperforms baseline methods, including UniAD and VAD, and exceeds the performance of distillation-based approaches such as ThinkTwice and DriveAdapter. In multi-ability evaluations, DiffAD demonstrates notable advantages over UniAD and VAD in interactive scenarios like merging and emergency braking. This improvement is attributed to its integrated learning framework, which enables explicit interactions among task objectives, resulting in more coherent and effective planning. 
Due to the relatively small size of the training dataset, DiffAD exhibits slightly lower performance in Traffic Sign compared to methods utilizing expert feature distillation. Incorporating expert features, which provide valuable driving knowledge, could help mitigate potential overfitting. Consequently, models leveraging expert feature distillation (e.g., TCP, ThinkTwice, and DriveAdapter) generally outperform those without it (e.g., VAD and UniAD).

\begin{figure}[h]
    
    \centering
    \begin{tikzpicture}
        \pie[
            text=legend,
            radius=2.5, 
            color={green!50!lime, red!70!yellow, yellow, orange!90, cyan}
        ]{
            38.64/SUCCESS (85),
            44.54/COLLISION (98),
            7.3/RED LIGHT (16),
            7.3/TIMEOUT (16),
            2.27/YIELD (5)
        }
    \end{tikzpicture}
    \caption{Status Distribution.}
    \label{fig:stas_pie}
\end{figure}
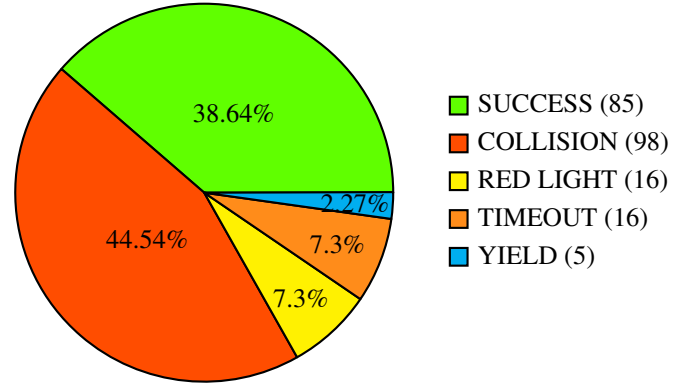

We conducted a failure case analysis of DiffAD, as shown in \cref{fig:stas_pie}. The analysis reveals that a significant portion of route failures was caused by collisions with traffic agents, indicating the challenges of interacting in CARLA v2. Additionally, a small number of failures were attributed to timeouts, typically caused by the planning module's occasional inability to resume motion after stopping, this issue can be effectively mitigated by utilizing expert distillation or incorporating more traffic light interactions data. A small percentage of failures occurred when the agent ran a red light, likely due to the low-quality rendering of traffic lights in CARLA or challenging lighting conditions, making them difficult to detect.

\subsection{Ablation Study}

\paragraph{The impact of denoising steps on performance}
DiffAD's iterative denoising follows a coarse-to-fine refinement process, progressively improving perception and planning. As shown in \cref{tab:desnoising_steps}, increasing denoising steps from 3 to 10 significantly reduces FID (-53.5\%) while improving Driving Score (+2.18) and Success Rate (+3.64), demonstrating that multi-step refinement helps resolve trajectory ambiguities. However, extending steps beyond 10 (e.g., to 20) leads to performance saturation, suggesting an optimal balance between computational overhead and planning precision. 

\begin{table}[t]
\centering
\caption{The impact of denoising steps on performance}
\label{tab:desnoising_steps}
\begin{tabular}{ccccc}
    \toprule
    Steps & FID $\downarrow$ & Driving Score $\uparrow$ & Success Rate $\uparrow$ \\ \midrule
    3 & 78.19 & 64.78 & 33.63 \\
    5 & 50.80 & 65.78 & 34.55 \\
    10 & 46.90 & \textbf{66.96} & \textbf{37.27} \\
    20 & \textbf{45.09} & 66.42 & 35.91 \\
    \bottomrule
\end{tabular}
\end{table}

\paragraph{The impact of tasks joint optimization}
We investigate the impact of jointly optimizing auxiliary tasks on planning performance. As shown in \cref{tab:auxiliary_task}, the full joint optimization of all three tasks achieves the best results, highlighting the importance of task joint optimization in enhancing planning performance.

\begin{table}[htbp]
\setlength{\tabcolsep}{1.5pt}
\centering
\caption{The impact of joint optimization of auxiliary tasks}
\label{tab:auxiliary_task}
\begin{tabular}{cccccccc}
\toprule
Detection & Motion & Planning & Driving Score $\uparrow$ & Success Rate $\uparrow$ \\
\midrule
/ & / & \checkmark & 30.11 & 5.9 \\
\checkmark & / & \checkmark & 59.10 & 29.09 \\
\checkmark & \checkmark & \checkmark & \textbf{66.96} & \textbf{37.27} \\
\bottomrule
\end{tabular}

\end{table}

\paragraph{The impact of Action Guidance dropout.}
Intuitively, relying too heavily on previous decisions can increase response latency in critical emergency scenarios. Conversely, making decisions without considering prior actions can lead to abrupt perception errors, resulting in unrealistic planning outcomes. To better understand this trade-off, we analyze the impact of different dropout rates in the Action Guidance module on planning performance.  \cref{tab:dropout} presents the results for various dropout rates. A dropout rate of 0.95 achieves the best balance, indicating that retaining a small portion of previous action guidance is beneficial for robust planning.

\begin{table}[h]
\centering
\begin{tabular}{lccc}
\toprule
Drop Rate & FID$\downarrow$ & Driving Score $\uparrow$ & Success Rate $\uparrow$ \\
\midrule
0.5 & 47.37 & 66.28 & 35.91\\
0.75 & 47.22 & 66.47 & 35.45\\
0.95 & 47.08 & \textbf{67.92} & \textbf{38.64}\\
1.0 & \textbf{46.90} & 66.96 & 37.27\\
\bottomrule
\end{tabular}
\caption{Impact of Action Guidance dropout on planning performance.}
\label{tab:dropout}
\end{table}

\paragraph{Efficiency of Unified Generative Modeling}
As shown in ~\cref{tab:runtime_comparison}, despite having a larger parameter size (545.6M vs. 58.1M for VAD), DiffAD achieves competitive latency (258ms vs. 140ms) and real-time FPS (3.9). This efficiency is attributed to two key innovations:
\begin{itemize}
\item \textbf{Task-agnostic compression}: The VAE effectively compresses BEV image while preserving critical information, significantly reduce the number of tokens for interactions and refinements in transformer layers.

\item \textbf{Parallelized diffusion head}: Unlike sequential multitask pipelines, DiffAD employs a shared denoising network to optimize all driving tasks jointly, eliminating inefficiencies of cascaded inference.
\end{itemize}

With TensorRT-FP16, DiffAD achieves 23.8 FPS (42ms inference, \cref{tab:module_runtime}). Notably, 83\% of its runtime is dominated by the diffusion process, which can be further optimized via distillation techniques \cite{cm, rectified_flow} with minimal performance trade-offs.

\begin{table}[t]
\centering
\caption{Comparison of parameters and FPS on GeForce RTX 4090}
\label{tab:runtime_comparison}
\begin{tabular}{cccc}
    \toprule
    Model & Parameters(M) & Latency(ms) & FPS ($\downarrow$) \\ \midrule
    UniAD-base & 84.2 & 355 & 2.8\\
    VAD-base & 58.1 & \textbf{140} & \textbf{7.1} \\
    DiffAD-10steps & 545.6 & 258 & 3.9\\
    \bottomrule
\end{tabular}

\end{table}

\begin{table}[t]
\centering
\caption{Module runtime (ms) of DiffAD on GeForce RTX 4090.}
\label{tab:module_runtime}
\begin{tabular}{*{6}{c}}
\toprule
Model & BEV & DiT (10 steps) & TEN & Total & FPS \\ \midrule
FP32 & 41 & 214 & 3 & 258 & 3.9 \\
TRT-FP16 & 1.6 & 40 & 1 & 42 & 23.8 \\ \bottomrule
\end{tabular}

\end{table}

\begin{figure*}[ht]
    \centering
    
    \textbf{BEV GT \hspace{3.cm} \textcolor{blue}{Expert} vs  \textcolor{red}{Planning}  \hspace{3.cm}  BEV Pred}
    
    \vspace{0.2cm} %
    
    \begin{subfigure}[b]{0.8\textwidth}
    \includegraphics[width=\textwidth]{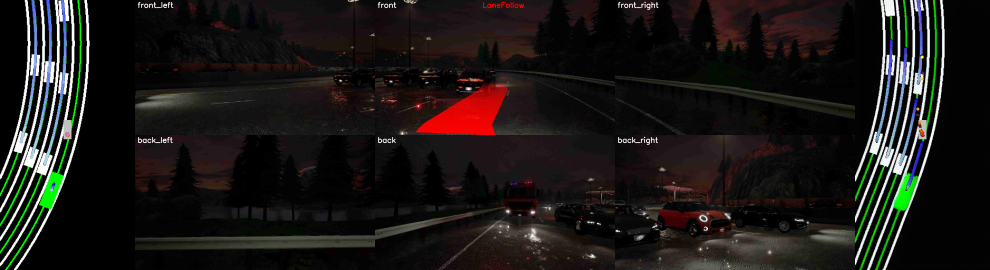}

    \includegraphics[width=\textwidth]{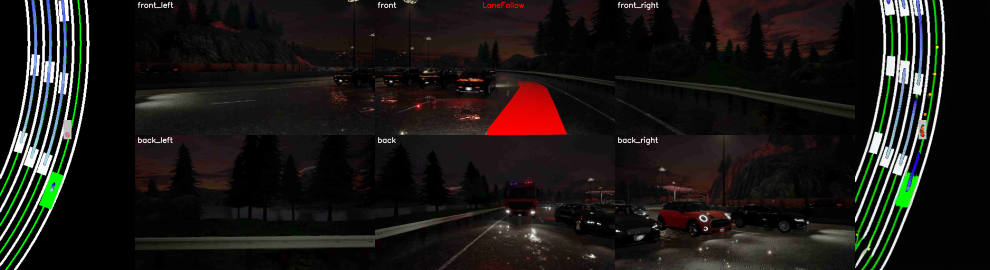}
    \caption{
    \textbf{Yielding to an Emergency Vehicle:} In the top image, the \textcolor{gray}{ego} vehicle changes lanes to the left and gradually merges with traffic to yield to an approaching \textcolor{green}{emergency vehicle}. In the bottom image, the \textcolor{gray}{ego} vehicle cancels the lane change and returns to its original lane due to heavy traffic.}
    
    \label{fig:b2d_nudge}
    \end{subfigure}

    \vspace{0.2cm} %
    
    \begin{subfigure}[b]{0.8\textwidth}
    \includegraphics[width=\textwidth]{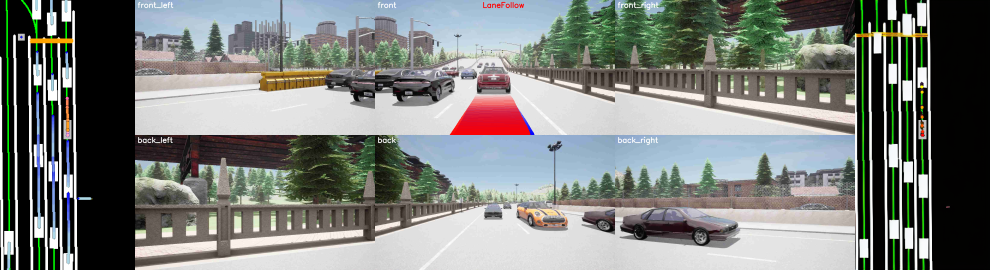}
    \includegraphics[width=\textwidth]{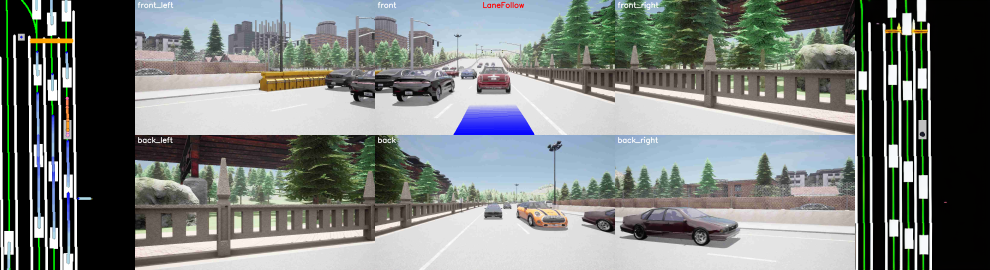}

    \caption{\textbf{T-Junction:} In the top image, the ego vehicle moving slowing to follow the lead vehicle. In the bottom image, the ego vehicle comes to a stop, maintaining a safe distance as the lead vehicle slowly approaches the stop line.}
    \label{fig:b2d_nudge}
    \end{subfigure}

    \caption{Demonstration of the Model's Multi-Modal Decision-Making}
    \label{fig:multi-modal}
\end{figure*}

\paragraph{The Multi-modality of Generative Modeling}
In \cref{fig:multi-modal}, we present qualitative results that showcase DiffAD’s powerful generative capabilities and its ability to produce diverse planning outcomes. For each scenario, we generate two decisions by sampling different latent variables. To enhance clarity, we overlay planned trajectories (in \textcolor{red}{red}) and expert trajectories (in \textcolor{blue}{blue}) onto the raw front-view image from surrounding cameras. The BEV ground truth (GT) is displayed on the left, while the predicted BEV is shown on the right. Notably, the generated BEV closely aligns with the ground truth, and the diverse planned trajectories is consistently safe and reasonable. This demonstrates DiffAD’s ability to accurately perceive the environment and effectively learn interactive behaviors.

\section{Conclusion and Future work}
\label{sec:conclusion} 
In this work, we present DiffAD, an end-to-end autonomous driving model built on a diffusion-based framework. Our key contribution lies in transforming heterogeneous targets of driving tasks into a unified rasterized representation, framing the E2E-AD as a conditional image generation task. This approach simplifies the problem and provides a clear pathway for leveraging various generative models, such as Diffusion models, GANs, VAEs, and auto-regressive models. We believe the strong performance of DiffAD highlights the potential of generative models in advancing autonomous driving research and hope it inspires further exploration in the field.

\paragraph{Limitations and future work.} 
Despite promising, the success rate on Carla v2 is still far from perfect. Effectively leveraging multi-modality generative predictions for planning, as well as aligning the model outputs with human preferences, are worth for further exploration. Additionally, there is a significant gap between the traffic simulation in Carla and real-world conditions. To address this, we are working towards deploying the system onboard to evaluate its performance in real traffic scenarios.

{
    \small
    \bibliographystyle{ieeenat_fullname}
    \bibliography{main}
}

\clearpage
\setcounter{page}{1}
\maketitlesupplementary

\section{Additional Experiments}

\label{sec:rationale}

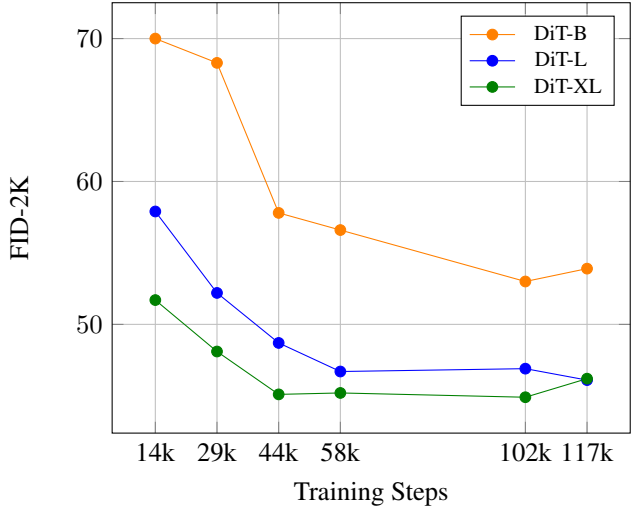
\begin{figure}[ht]
    \centering
    \begin{tikzpicture}
    \begin{axis}[
        xlabel={Training Steps},
        ylabel={FID-2K},
        grid=both,
        xtick={1,2,3,4,7,8},
        xticklabels={14k,29k,44k,58k,102k,117k}, 
        legend pos=north east,
        legend style={nodes={scale=0.8, transform shape}},
        ]
        \addplot[color=orange,mark=*]
            coordinates {
                (1,70.0) (2,68.3) (3,57.8) (4,56.6)
                (7, 53.0) (8, 53.9)
            };
        \addlegendentry{DiT-B}
    
        \addplot[color=blue,mark=*]
            coordinates {
                (1,57.9) (2,52.2) (3, 48.7) (4, 46.7) (7, 46.9) (8, 46.1)
            };
        \addlegendentry{DiT-L}
    
        \addplot[color=green!50!black,mark=*]
            coordinates {
                (1, 51.7) (2, 48.1) (3, 45.1) (4, 45.2) (7, 44.9) (8, 46.2)
            };
        \addlegendentry{\hspace{1.5mm} DiT-XL}

    \end{axis}
    \end{tikzpicture}
    \caption{Comparison of different DiT configs.}
    \label{fig:scaling_up}

\end{figure}

\paragraph{Scaling up parameters.} We investigate how increasing model parameters enhances capacity by training three DiffAD models with different DiT configurations (B, L, XL), as detailed in \cref{tab:dit_cfg}. Throughout training of temporal stage, we track FID-2K scores, as illustrated in \cref{fig:scaling_up}, which show a consistent improvement in FID as the transformer depth and width increase. Since configurations L and XL achieve similar FID scores, we adopt DiT-L as the default setting for experiments.

\begin{table}[h]
\centering
\begin{tabular}{lcccc}
\toprule
Model & Layers $N$ & Hidden size $d$ & Heads & Params(M) \\
\midrule
DiT-B & 12 & 768 & 12 & 131.8 \\
DiT-L & 24 & 1024 & 16 & 460.3 \\
DiT-XL & 28 & 1152 & 16 & 677.9 \\
\bottomrule
\end{tabular}
\caption{Details of DiT models.}
\label{tab:dit_cfg}
\end{table}

\begin{figure}[h]
    \centering
    \begin{subfigure}{0.09\textwidth}
        \centering
        \textcolor{blue}{NFE=3}
    \end{subfigure}
    \begin{subfigure}{0.09\textwidth}
        \centering
        \textcolor{green}{NFE=10}
    \end{subfigure}
    \begin{subfigure}{0.09\textwidth}
        \centering
        \textcolor{red}{NFE=20}
    \end{subfigure}
    \begin{subfigure}{0.09\textwidth}
        \centering
        \textcolor{purple}{NFE=100}
    \end{subfigure}
    \begin{subfigure}{0.09\textwidth}
        \centering
        \textcolor{black}{GT}
    \end{subfigure}

    \begin{subfigure}{0.09\textwidth}
        \centering
        \includegraphics[width=\linewidth]{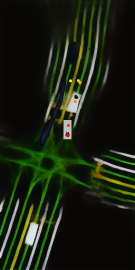}
    \end{subfigure}
    \begin{subfigure}{0.09\textwidth}
        \centering
        \includegraphics[width=\linewidth]{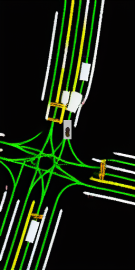}
    \end{subfigure}
    \begin{subfigure}{0.09\textwidth}
        \centering
        \includegraphics[width=\linewidth]{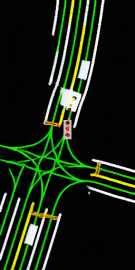}
    \end{subfigure}
    \begin{subfigure}{0.09\textwidth}
        \centering
        \includegraphics[width=\linewidth]{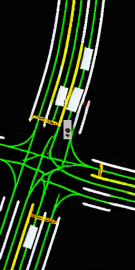}
    \end{subfigure}
    \begin{subfigure}{0.09\textwidth}
        \centering
        \includegraphics[width=\linewidth]{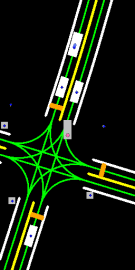}
    \end{subfigure}
    
     \begin{subfigure}{0.09\textwidth}
        \centering
        \includegraphics[width=\linewidth]{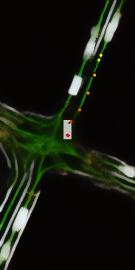}
    \end{subfigure}
    \begin{subfigure}{0.09\textwidth}
        \centering
        \includegraphics[width=\linewidth]{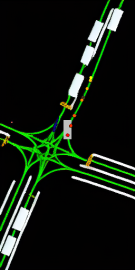}
    \end{subfigure}
    \begin{subfigure}{0.09\textwidth}
        \centering
        \includegraphics[width=\linewidth]{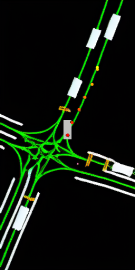}
    \end{subfigure}
    \begin{subfigure}{0.09\textwidth}
        \centering
        \includegraphics[width=\linewidth]{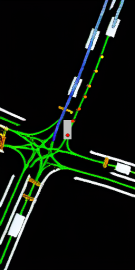}
    \end{subfigure}
    \begin{subfigure}{0.09\textwidth}
        \centering
        \includegraphics[width=\linewidth]{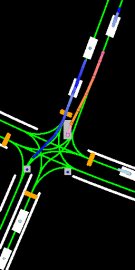}
    \end{subfigure}

     \begin{subfigure}{0.09\textwidth}
        \centering
        \includegraphics[width=\linewidth]{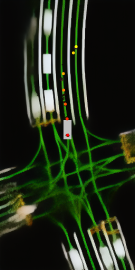}
    \end{subfigure}
    \begin{subfigure}{0.09\textwidth}
        \centering
        \includegraphics[width=\linewidth]{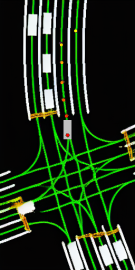}
    \end{subfigure}
    \begin{subfigure}{0.09\textwidth}
        \centering
        \includegraphics[width=\linewidth]{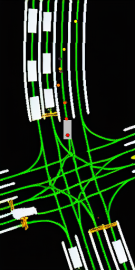}
    \end{subfigure}
    \begin{subfigure}{0.09\textwidth}
        \centering
        \includegraphics[width=\linewidth]{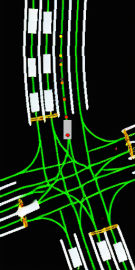}
    \end{subfigure}
    \begin{subfigure}{0.09\textwidth}
        \centering
        \includegraphics[width=\linewidth]{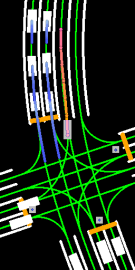}
    \end{subfigure}

    \caption{Effect of different NFE values on image quality}
    \label{fig:nfe}
\end{figure}

\paragraph{Visualization of different denoising steps}
As demonstrated in earlier experiments, increasing the number of denoising steps improves both image quality and planning performance. However, excessively high NFE introduces additional computational costs without yielding significant planning benefits. To further investigate the impact of NFE, we visualize the generated BEV images at different denoising steps. As shown in \cref{fig:nfe}, increasing NFE enhances the clarity of traffic elements. When NFE=10, the system successfully captures sufficient environmental details, ensuring reliable planning performance. Further denoising primarily refines minor details of traffic elements without substantially affecting planning outcomes. This experiment highlights that placing excessive emphasis on perception tasks may not necessarily lead to meaningful improvements in planning performance, suggesting the need for a balanced allocation of computational resources between perception and planning.

\paragraph{Open-loop metric vs Closed-loop metric} As reported in Bench2Drive, open-loop evaluation ignores critical factors such as distribution shift and causal confusion, thus it fails to assess a model’s ability to handle dynamic interactions. As shown in \cref{tab:comparison_open_closed}, DiffAD achieves the best closed-loop performance in terms of Driving Score and Success Rate, despite having a higher L2 error than UniAD and DriveAdapter. This highlights the importance of closed-loop evaluation in capturing the true effectiveness of autonomous driving models.

\begin{table}[t]
\setlength{\tabcolsep}{2pt}
\centering
\caption{Comparison of Open-loop and Closed-loop metrics on Bench2Drive.}
\label{tab:comparison_open_closed}
\begin{tabular}{lccc}
\toprule
    Method & Avg. L2 $\downarrow$ & Driving Score $\uparrow$ & Success Rate $\uparrow$ \\ \midrule
    UniAD-Base & \textbf{0.73} & 45.81 & 16.36 \\
    TCP-traj* & 1.70 & 59.90 & 30.00 \\
    DriveAdapter* & 1.01 & 64.22 & 33.08 \\
    DiffAD & 1.55 & \textbf{67.92} & \textbf{38.64} \\
\bottomrule
\end{tabular}

\end{table}
\end{document}